\documentclass{article}

\usepackage{microtype}
\usepackage{graphicx}
\usepackage{subcaption}
\usepackage{booktabs} 

\usepackage{hyperref}
\usepackage{url}
\usepackage[utf8]{inputenc} 
\usepackage[T1]{fontenc}    
\usepackage{amsmath,bm}
\usepackage{xr-hyper} 
\usepackage{amsfonts}       
\usepackage{nicefrac}       
\usepackage{xcolor}         
\usepackage{amsthm}
\usepackage{comment}
\usepackage{makecell}
\usepackage{thm-restate}
\usepackage{soul}
\usepackage[normalem]{ulem}
\usepackage{caption}
\usepackage{dashrule}

\definecolor{BetterBlue}{rgb}{0.38, 0.44, 1.0}
\definecolor{OSUOrange}{rgb}{1,0.3,0.5}

\newcommand{\benchmarkname}{Noisy-Diff }

\usepackage{hyperref}

\usepackage[accepted]{icml2026}

\usepackage{amsmath}
\usepackage{amssymb}
\usepackage{mathtools}
\usepackage{amsthm}
\usepackage{stfloats}
\usepackage{cuted}

\usepackage[capitalize,noabbrev]{cleveref}

\usepackage[textsize=tiny]{todonotes}

\icmltitlerunning{LatentDiff: Scaling Semantic Dataset Comparison to Millions of Images}

\begin{document}

\twocolumn[
  \icmltitle{LatentDiff: Scaling Semantic Dataset Comparison to Millions of Images}



  \icmlsetsymbol{equal}{*}
  \begin{icmlauthorlist}
    \icmlauthor{James Flora}{yyy}
    \icmlauthor{Kowshik Thopalli}{comp}
    \icmlauthor{Akshay R. Kulkarni}{zzz}
    \icmlauthor{Weng-Keen Wong}{yyy}
    \icmlauthor{Shusen Liu}{comp}
  \end{icmlauthorlist}

  \icmlaffiliation{yyy}{School of Electrical Engineering and Computer Science, Oregon State University, Corvallis, Oregon, USA}
  \icmlaffiliation{comp}{Lawrence Livermore National Laboratory, Livermore, California, USA}
  \icmlaffiliation{zzz}{University of California San Diego, California, USA}

  \icmlcorrespondingauthor{James Flora}{floraj@oregonstate.edu}

  \icmlkeywords{Machine Learning, ICML}

  \vspace{0.2in}
  \centerline{\includegraphics[width=0.75\textwidth]{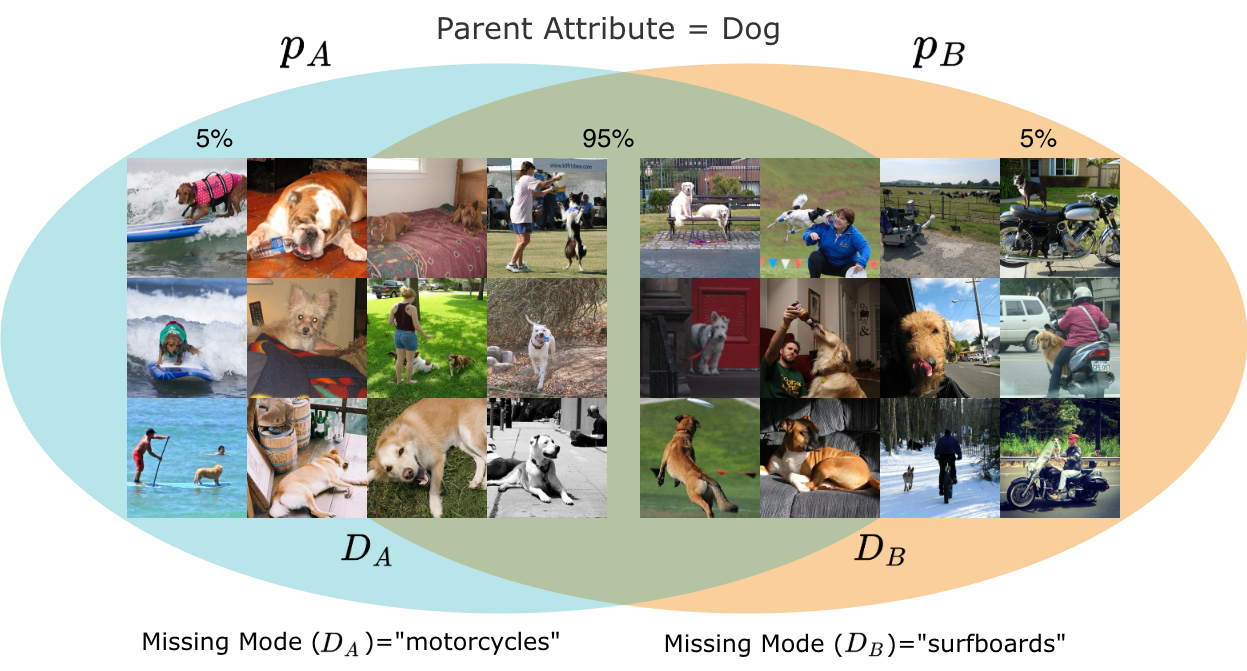}}

  {\small
\textbf{Figure 1: Semantic Comparison of Noisy Datasets.}
Construction of naturally noisy datasets for which missing modes make up $\sim5\%$ of the total data. 
A successful output labels these missing modes with text. Though $D_A$ and $D_B$ both contain pictures of dogs, $D_A$ does not contain any pictures of dogs and motorcycles, while $D_B$ does not contain any pictures of dogs and surfboards.
}
    \refstepcounter{figure}
  \label{fig:venn}
  \vspace{0.2in}
]



\makeatletter
\renewcommand{\Notice@String}{}
\makeatother
\printAffiliationsAndNotice{}  
\setcounter{figure}{1}
\begin{abstract}
We present \texttt{LatentDiff}, a scalable framework for semantic dataset comparison that operates directly in the latent space of pretrained vision encoders. By combining sparse autoencoder-based divergence testing with density ratio estimation, \texttt{LatentDiff} identifies interpretable semantic differences between datasets at a fraction of the computational cost of caption-based alternatives. We also introduce \texttt{Noisy-Diff}, a benchmark capturing realistic sparse distribution shifts that cause existing methods to struggle. Experiments demonstrate that \texttt{LatentDiff} achieves superior accuracy while remaining robust to settings where an extremely small fraction of images  (from 5\% to <1\% ) differ semantically. 
\end{abstract}
\section{Introduction}

Comparing image datasets is fundamental to many critical tasks in machine learning \citep{koh2021wilds, kulinski2023explaining, torralba2011unbiased}. The core question is simple: given two sets of images, what semantic concepts distinguish them? Despite this simplicity, existing approaches face two key challenges: (i) methods struggle to provide interpretable explanations while remaining scalable, and (ii) benchmarks assume cleanly separated distributions, whereas real-world differences are often subtle and sparse.

Regarding the first challenge, current methods face a tradeoff between interpretability and scalability. Caption-based approaches~\citep{dunlap2024visdiff} produce human-readable descriptions but require expensive per-image inference. Statistical tests on pretrained embeddings scale efficiently but reveal little about \emph{how} distributions differ. Regarding the second challenge, existing benchmarks~\citep{dunlap2024visdiff} feature pervasive differences across datasets. Real-world shifts, however, are often subtle and affect only a small subset, e.g., generative model failures on certain categories.

To address the scalability gap, we formulate a semantic dataset comparison directly in the image latent space using sparse autoencoders (SAEs). Recently shown to transfer effectively to vision encoders~\citep{pach2025sparseautoencoderslearnmonosemantic, zaigrajew2025interpreting}, SAEs decompose embeddings into sparse, approximately monosemantic features that often align with interpretable visual concepts. By mapping each image onto a fixed dictionary of attributes, we compare activation distributions over SAE features rather than raw images or opaque embeddings, identifying which features are over- or under-represented in one dataset relative to another.
However, SAE-based analysis is limited by its learned vocabulary. As a result, differences involving attributes outside the dictionary may go undetected. To complement it, we introduce a lightweight mechanism based on density ratio estimation (DRE) that localizes where datasets diverge most, rather than explaining differences directly. Operating in a continuous latent space, DRE identifies highly discriminative samples, enabling targeted application of expensive captioning and language model reasoning for open-ended discovery beyond the SAE's vocabulary.

Given these complementary viewpoints, we propose a simple ensembling strategy. SAE-based hypotheses provide interpretable explanations when relevant attributes lie within the learned dictionary, while DRE-based hypotheses localize salient differences outside this vocabulary. We aggregate top-ranked hypotheses from each branch into a joint candidate set, which empirically improves coverage of valid semantic differences.

Finally, to address the benchmark gap, we introduce a more challenging benchmark with controlled, sparse semantic differences and distractors. Unlike existing benchmarks, where nearly all images differ between sets, ours embeds subtle category-level shifts affecting only a small fraction of each dataset. This \emph{needle-in-a-haystack} setting exposes the brittleness of methods that rely on random sampling and aggregate reasoning. Our contributions are threefold:

\begin{itemize}
\item We propose \texttt{LatentDiff}, a scalable framework for interpretable dataset comparison in pretrained image encoders' latent space.
\item We introduce \texttt{Noisy-Diff}, a benchmark capturing realistic, sparse distribution shifts that provides a rigorous testbed for semantic difference methods.
\item We demonstrate that \texttt{LatentDiff} can significantly outperform caption-based alternatives at a fraction of the computation cost.
\end{itemize}
\section{Related Work}
\label{sec:related_work}




Our work sits at the intersection of semantic dataset comparison, interpretability via sparse representations, and distributional analysis in embedding spaces.

\textbf{Dataset Comparison with Natural Language.}
\citet{zhong2022d3} pioneered automatic dataset difference description for text using fine-tuned GPT-3, achieving 76\% human agreement, and their follow-up work~\citep{zhong2023d5} extended this to goal-driven discovery. For images, VisDiff~\citep{dunlap2024visdiff} introduced set difference captioning using BLIP-2 captions with GPT-4, achieving 61\% accuracy on hard cases but requiring expensive per-image captioning. GSCLIP~\citep{zhu2022gsclip} provides a training-free framework with predefined text banks. Domino~\citep{eyuboglu2022domino} discovers model failure slices using cross-modal CLIP embeddings. \citet{olson2021unsupervised} introduced unsupervised attribute alignment for distribution shifts, with Cross-GAN Auditing~\citep{olson2023cross} extending this to generative model comparison. Unlike these approaches, we exploit geometric structure in pretrained vision encoder spaces directly, enabling efficient large-scale comparison through learned sparse features.

\textbf{Distribution Shift Detection Benchmarks and Tools.}
The WILDS benchmark~\citep{koh2021wilds} curates datasets reflecting real-world shifts. The ImageNet robustness suite, i.e., ImageNet-V2~\citep{recht2019imagenetv2}, ImageNet-C~\citep{hendrycks2019imagenetc}, ImageNet-A~\citep{hendrycks2021imageneta}, ImageNet-R~\citep{hendrycks2021imagenetr}) evaluates model robustness under various distribution shifts. \citet{taori2020measuring} found that synthetic robustness rarely transfers to natural shifts. \citet{rabanser2019failing} investigated shift detection combining dimensionality reduction with two-sample testing. These benchmarks assume pervasive differences; we address sparse differences where only small sample fractions differ.
MetaShift~\citep{liang2022metashift} creates image sets for studying contextual shifts, and serves as inspiration for our own \texttt{NoisyDiff}. 
There are also interactive methods for understanding data shifts, including VLSlice~\citep{slyman2023vlslice}, FACTS~\citep{yenamandra2023facts}, and LADDER~\citep{ghosh2025ladder}.

\textbf{Sparse Autoencoders for Interpretable Features.}
\citet{bricken2023monosemanticity} demonstrated SAEs can decompose transformers into thousands of interpretable features. \citet{cunningham2023sparse} showed SAE features are more monosemantic than neurons. \citet{gao2024scaling} developed a scaling methodology for extremely wide SAEs. For vision, \citet{templeton2024scaling} scaled SAE to Claude 3 Sonnet, while \citet{joseph2025steering} applied SAEs to CLIP vision transformers. We leverage these advances to decompose CLIP embeddings into concept-level features for divergence-based comparison.

\textbf{Density Ratio Estimation.}
Density ratio estimation (DRE) enables distribution comparison without explicit density estimation. Foundational methods include KLIEP~\citep{sugiyama2007kliep}, uLSIF~\citep{kanamori2009lsif}, and RuLSIF~\citep{yamada2013rulsif}. \citep{sugiyama2012density} provides a comprehensive theoretical foundation on DRE methods. DRE has been applied to various techniques, e.g., change-point detection~\citep{liu2013changepoint}, outlier detection~\citep{hido2011outlier}, and covariate shift adaptation~\citep{sugiyama2007covariate}. 
We employ DRE in pretrained vision encoder latent spaces to identify maximally discriminative samples.


\section{Problem Statement}

We first formalize the task of semantic dataset comparison. Given two image datasets $D_A$ and $D_B$, our goal is to identify and describe the \emph{missing modes} of each dataset relative to the other. A missing mode refers to a semantic pattern, visual concept, or scene configuration that appears reliably in one dataset but is absent or underrepresented in the other. For example, if $D_A$ contains images of dogs on surfboards while $D_B$ does not, then ``surfboards'' constitutes a missing mode of $D_B$ relative to $D_A$.

\paragraph{Formal Setup.} Let $\mathcal{X}$ denote the space of natural images and let $\mathcal{C}$ denote a universe of semantic concepts (e.g., object categories, visual attributes, scene types). We are given two finite datasets $D_A, D_B \subset \mathcal{X}$ drawn from underlying distributions $p_A$ and $p_B$, respectively (see Figure \ref{fig:venn}). Each image $x \in \mathcal{X}$ is associated with a set of concepts $c(x) \subseteq \mathcal{C}$ that describe its semantic content.

We define the \emph{missing modes of $D_B$ relative to $D_A$}, denoted $M_{A \to B}$, as the set of concepts that are substantially more prevalent in $D_A$ than in $D_B$:
\begin{equation}
    M_{A \to B} = \left\{ c \in \mathcal{C} \;\middle|\; \Pr_{x \sim p_A}[c \in c(x)] \gg \Pr_{x \sim p_B}[c \in c(x)] \right\}.
\end{equation}
Symmetrically, the \emph{missing modes of $D_A$ relative to $D_B$} are defined as:
\begin{equation}
    M_{B \to A} = \left\{ c \in \mathcal{C} \;\middle|\; \Pr_{x \sim p_B}[c \in c(x)] \gg \Pr_{x \sim p_A}[c \in c(x)] \right\}.
\end{equation}

The objective of semantic dataset comparison is to recover $M_{A \to B}$ and $M_{B \to A}$ from finite samples $D_A$ and $D_B$.

\paragraph{Desired Output.} The output of a semantic dataset comparison method consists of natural language descriptions characterizing the semantic differences between $D_A$ and $D_B$, specifying which concepts are under represented in each dataset with respect to the other.

\begin{figure*}[t]
  \centering
  \includegraphics[width=\textwidth]{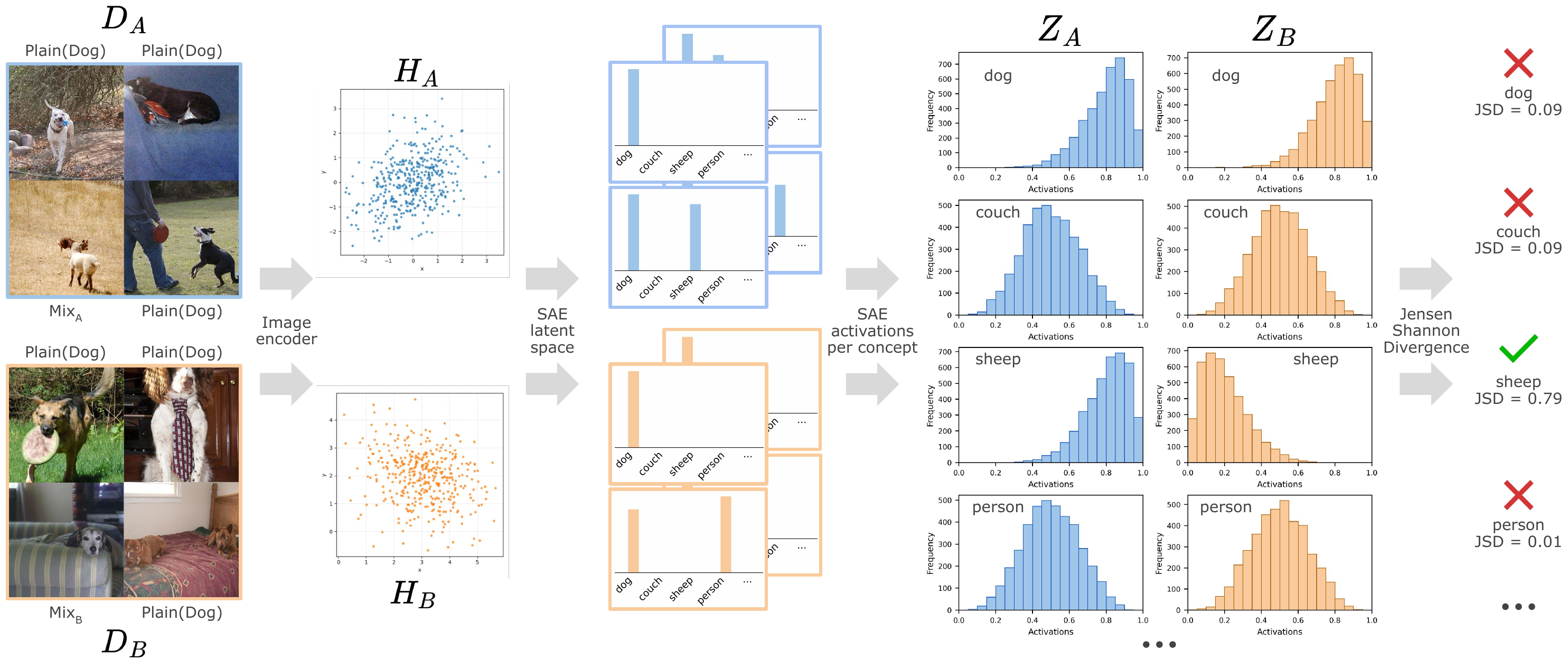}
  \caption{Workflow of \texttt{LatentDiff} (SAE). Images are embedded, projected into an SAE latent space, and candidate missing modes are identified by comparing neuron activation distributions via Jensen-Shannon Divergence.}
  \label{fig:workflow}
\end{figure*}
\section{LatentDiff}

In this section, we present \texttt{LatentDiff}, a latent representation approach to semantic image dataset comparison.
Modern vision encoders \citep{radford2021clip, zhangdino} learn latent spaces where geometric proximity reflects semantic similarity. For CLIP \citep{radford2021clip}, this structure emerges from large scale contrastive pretraining and encodes rich visual concepts that transfer across tasks. By operating directly in this space, we can leverage standard statistical tools for distribution comparison without expensive per image processing. The key insight is that pretrained encoders have already extracted visual semantics; our task is to quantify the distributional differences this representation reveals and link them to interpretable labels.
We employ two complementary strategies. The first identifies disentangled concepts within the latent space and performs per dimension comparison to surface interpretable differences directly. The second locates global distributional divergence in the continuous space and then interprets semantic differences from the most extreme samples.

These two strategies are complementary. The SAE based approach efficiently identifies multiple axes of difference in a single pass but is inherently limited to concepts within the SAE's feature vocabulary. The DRE based approach operates over continuous latent spaces and can reveal differences involving novel or fine grained concepts that may not correspond to any SAE feature. Together, they provide both breadth of coverage and robustness to vocabulary gaps. We discuss necessary the necessary preliminaries in \Cref{apx:background}


\subsection{ Concepts Shift Identification with SAE Activations}
For interpretable concept identification, we leverage sparse autoencoders that decompose pretrained vision encoder representations into disentangled, approximately monosemantic features. 
By treating the activations of each SAE neuron (concept dimension) as separate distributions over datasets $A$ and $B$, we apply per-neuron divergence tests to extract overrepresented concepts in one set relative to the other. 
Concretely, an SAE learns an overcomplete sparse representation of latent features produced by a pretrained vision encoder by minimizing a reconstruction--sparsity tradeoff (see Appendix~\ref{appendix:SAE}).

Let $H_A$ and $H_B$ be the latent space point clouds of datasets $A$ and $B$, and let a trained SAE map any latent $h \in \mathbb{R}^d$ to a sparse code,
\[
z = g(h) \in \mathbb{R}^k
\] where $g : \mathbb{R}^d \rightarrow \mathbb{R}^k$ denotes the SAE encoder and $k$ denotes the number of SAE neurons (dictionary elements), typically with $k \gg d$.
Applying $g$ pointwise yields two activation matrices
\[
Z_A = [z_i^{(A)}]_{i=1}^{n_A} \in \mathbb{R}^{n_A \times k},\qquad
Z_B = [z_i^{(B)}]_{i=1}^{n_B} \in \mathbb{R}^{n_B \times k}
\]

We write $Z_{A[:,j]}$ for column $j$ (corresponding to neuron $j$ in the SAE, also referred to as concept $j$) in $Z_A$, and likewise for $Z_B$. For each concept $j$, we compare its activation distributions in $A$ vs. $B$.

Let $\pi_j^A$ and $\pi_j^B$ be empirical one-dimensional densities of $Z_{A[:,j]}$ and $Z_{B[:,j]}$. We score the difference between these two distributions with Jensen-Shannon divergence
\[
JSD(\pi_j^A, \pi_j^B) = \frac{1}{2}D_{KL}(\pi_j^A || \eta_j) + \frac{1}{2}D_{KL}(\pi_j^B || \eta_j)
\]
where $D_{KL}$ is the KL divergence and $\eta_j = \frac{1}{2}(\pi_j^A + \pi_j^B)$ is a mixture distribution. We then compute the signed mean gap,
$\Delta \mu_j = \mathbb{E}[Z_{A[:,j]}] - \mathbb{E}[Z_{B[:,j]}]$ and then rank the concepts (neurons) by JSD and use sign$(\Delta \mu_j)$ to assign directionality.
\begin{itemize}
    \item $A$-biased concepts (more present in $D_A$): $\Delta \mu_j > 0$, sorted by JSD
    \item $B$-biased concepts (more present in $D_B$): $\Delta \mu_j < 0$ sorted by JSD
\end{itemize}

This provides a list of concepts from the SAE that are said to be ``more representative'' of $D_A$ and vice versa, with the entire workflow shown in \Cref{fig:workflow}.  In practice, the one-dimensional empirical densities are estimated using histogram-based discretization using the Freedman-Diaconis rule. We discuss further pruning strategies and the assignment of human-interpretable labels to SAE concepts in Section~\ref{sec:experiments}. 


\subsection{Localization of Dataset Differences via Density Ratio Estimation}

 While SAE-based analysis adopts a \emph{disentangle/interpret-then-compare} strategy for semantic dataset comparison, an alternative and complementary perspective is to first identify where two datasets differ most in a suitable latent space, and then interpret those differences post hoc. We adopt this \emph{distinguish-then-interpret} view through density ratio estimation (DRE), which provides a scalable mechanism for localizing maximally discriminative samples between datasets. This localization-first perspective is particularly useful when salient differences are sparse, rare, or lie outside the fixed feature vocabulary of the sparse autoencoders.

Concretely, we train a lightweight density ratio model in the encoder’s latent space to estimate how characteristic each sample is of one dataset relative to the other. Let $f_{\theta}: \mathbb{R}^d \rightarrow \mathbb{R}$ denote a learned ratio head trained on latent representations $H_A$ and ${H}_B$ as a probabilistic classifier:
\[
p_{\theta}(y = 1 \mid h) = \sigma(f_{\theta}(h)), \quad y = 1\ \text{for}\ A,\; y = 0\ \text{for}\ B,
\]
which approximates the log density ratio $\log r(h)$ between the two distributions (see Section~\ref{subsec:dre_bg}).\\
 Ranking samples by $\log r(h)$ and selecting the top-$k$ from each side yields: (i) large positive $\log r(h)$: samples most characteristic of $D_A$; large negative $\log r(h)$: samples most characteristic of $D_B$.



As we demonstrate empirically, this localization strategy is particularly effective under noisy comparison regimes, where semantic differences are sparse and the two datasets exhibit substantial overlap in latent space. This robustness arises because density-ratio ranking depends on relative density contrasts rather than precise decision boundary geometry. As a result, even when the classifier-induced decision boundary separating the two latent distributions is poorly defined, the likelihood-ratio ordering remains reliable for identifying rare but salient missing modes.

Note that because this analysis operates directly on distributions of images in latent space, its output consists of characteristic image examples rather than explicit semantic descriptions. While such examples effectively localize where the datasets differ, interpreting these differences in a human-meaningful way requires two additional steps. First, we apply captioning to the DRE-selected images using a vision-language model. Second, we aggregate the resulting per-image captions with a large language model to generate natural language descriptions of the differences between the two datasets.  Although captioning and LLM-based aggregation are computationally expensive in general, here they are applied only to a small number of highly discriminative samples (typically fewer than ten), resulting in minimal overhead while avoiding the brittleness of global, sampling-based captioning approaches.

\subsection{Combined Complementary Hypothesis Generation}
Given the complementary viewpoints, strengths, and failure modes of the SAE- and DRE-based approaches, we combine their outputs to improve coverage of semantic differences. Rather than ensembling models in the traditional sense, we aggregate hypotheses produced by these two mechanisms, which operate on the same underlying latent representations.

In practice, we apply the SAE pipeline to obtain a ranked list of concept-level differences along interpretable, fixed semantic dimensions. In parallel, we use DRE to retrieve a small set of maximally discriminative samples from each dataset and derive open-vocabulary hypotheses by captioning these samples and aggregating the resulting captions with a large language model. We then form a unified candidate set by taking the union of the top-$p$ SAE-derived concepts and the top-$q$ DRE-derived hypotheses.

This aggregation incurs minimal additional computational cost, as both components operate on cached encoder embeddings and DRE-based captioning is applied only to a small number of selected samples. As we show empirically in Section~\ref{sec:experiments}, combining these complementary hypothesis generators provides robust performance when differences are rare, sparse, or semantically fine-grained.

\begin{table*}[ht!]
  \centering
  \caption{Results on \texttt{Noisy-Diff} comparing \texttt{LatentDiff} to the baseline VisDiff on a baseline of 15 pairs per parent category. Cosine similarity is averaged over pairs; $\rho(\text{scarcity})$ is the Pearson correlation between per-pair performance and the scarcity of the missing mode; more negative means performance degrades as the mode gets scarcer. Average refers to the arithmetic mean for cosine similarity and the Fisher pooled probability test for $\rho$(scarce).}
  \label{tab:comparison_results}
  \begin{tabular}{l c c c c}
    \toprule
    \multicolumn{1}{c}{Parent} &
    \multicolumn{2}{c}{LatentDiff (SAE) (\textit{Ours})} &
    \multicolumn{2}{c}{VisDiff} \\
    \cmidrule(lr){2-3}\cmidrule(lr){4-5}
    & Cosine sim & $\rho(\text{scarcity})$ {\scriptsize(↓ worse)} & Cosine sim & $\rho(\text{scarcity})$ {\scriptsize(↓ worse)} \\

    \midrule
    bicycle      & 0.469 $\pm$ 0.139 &  -0.185 & 0.295 $\pm$ 0.281 &  -0.370 \\
    backpack     & 0.587 $\pm$ 0.105 &  0.242 & 0.366 $\pm$ 0.314 &  -0.068 \\
    cat          & 0.687 $\pm$ 0.186 &  0.031 & 0.241 $\pm$ 0.197 & -0.428 \\
    dog          & 0.545 $\pm$ 0.146 & -0.269 & 0.263 $\pm$ 0.205 & -0.383 \\
    tv           & 0.509 $\pm$ 0.133 &  0.134 & 0.300 $\pm$ 0.138& 0.429 \\
    motorcycle   & 0.492 $\pm$ 0.096 & -0.033 & 0.320 $\pm$ 0.217& -0.493 \\
    \midrule
    \textbf{Average} & \textbf{0.548} & \textbf{-0.014} & \textbf{0.298} & \textbf{-0.230} \\
    \bottomrule
  \end{tabular}
\end{table*}


\section{\benchmarkname: A Challenging Benchmark for Sparse Semantic Dataset Comparison}
\label{sec:noisediff}

Existing benchmarks for semantic dataset comparison, such as those introduced in VisDiff ~\citep{dunlap2024visdiff}, typically construct dataset pairs in which semantic differences are pervasive: nearly every image in one set differs meaningfully from images in the other. While this design simplifies evaluation, it fails to reflect conditions encountered in practice, where distribution shifts are often subtle and sparse.
To address this gap, we propose \texttt{Noisy-Diff}, a systematic protocol for constructing challenging semantic dataset comparison tasks from datasets with rich semantic annotations, such as MS-COCO. The protocol enables the creation of large, noisy dataset pairs that exhibit sparse, controlled semantic differences, mimicking real-world distribution shifts.


Given a parent object class $P$ (e.g., "dog") and two attribute classes $A$ and $B$ (e.g., "surfboard", "motorcycle"), we form:
$
\begin{aligned}
\mathrm{Plain}(P) &:= \{\, x \mid x \text{ contains } P \text{ but neither } A \text{ nor } B \,\},\\
\mathrm{Mix}_A    &:= \{\, x \mid x \text{ contains } P \land A \text{ but not } B \,\},\\
\mathrm{Mix}_B    &:= \{\, x \mid x \text{ contains } P \land B \text{ but not } A \,\}.
\end{aligned}
$

and construct two datasets $D_A = \mathrm{Plain}_1(P) \cup \mathrm{Mix}_A$, $\qquad D_B = \mathrm{Plain}_2(P) \cup \mathrm{Mix}_B$ where Plain$_1$ and Plain$_2$ are a disjoint partition of Plain$(P)$ as to avoid overlap, and all items in $P \cap A \cap B$ are removed. For example, if $P = $"dog", $A = $"surfboard", and $B = $"motorcycle", then $P_A$ contains images of dogs in many different contexts, including surfboards, but not motorcycles. In contrast, $P_B$ contains images of dogs in many different contexts, including motorcycles, but not surfboards.
We apply this protocol to both MS-COCO~\citep{lin2015microsoftcococommonobjects} and the ImageNet training set~\citep{deng2009imagenet} (details in the Appendix~\ref{apx:imagenet}), yielding large-scale, noisy evaluation pairs with controlled sparsity.


For evaluation, we treat the attribute labels $(A,B)$ as reference descriptions of the expected semantic differences for a given pair $(P;A,B)$. Rather than enforcing a single ground-truth label, we compute cosine similarity between predicted descriptions and these reference terms in a text embedding space, acknowledging that semantically related concepts (e.g., ocean'' for surfboard'') may constitute valid alternative descriptions of the missing mode.
\section{Experiments}
\label{sec:experiments}
In this section we discuss our experimental settings, splits, and hyperparameters. We compare against the VisDiff methodology proposed in \citet{dunlap2024visdiff} as the current SotA, and provide experimentation on their own proposed dataset found in Appendix~\ref{apx:visdiff}, along with qualitiative results beyond object-level shifts. We provide further experimentation against GSCLIP \cite{zhu2022gsclip} in Appendix~\ref{apx:gsclip} as a known baseline for latent methods as well as to broaden the evaluation protocol beyond cosine similarity.

\subsection{Dataset} \label{subsec:dataset}
Following the \texttt{Noisy-Diff} protocol, for the COCO dataset, we choose 6 different parent categories $P$ and form splits for each pair $(A,B)$ that has at least 100 images in both Mix$_A$ and Mix$_B$. Depending on $(P,A,B)$, the "missing mode" mass ranges from roughly $5\%$ to $50\%$ of the parent set. For example, choosing $A = $"person" typically yields $\approx 50\%$ due to ubiquity in COCO, whereas $A = $"banana" often yields the minimum ($\approx 100$ images, $\sim 5\%$). This variability is intentional: it stress-tests recovery under highly unbalanced (rare) modes.

For each parent $P$, we enumerate the set of all eligible attributes $E$. Each experiment fixes a pair $(A,B), A \neq B$, \emph{i.e.,} the number of experiments for parent $P$ is defined by the number of pairwise comparisons that can be done, $\binom{|E|}{2}$. Generally, $ 20 < |E| < 30$  so for each parent class we have anywhere from $75$ to $500$ experiments. We discard parents with fewer than $2000$ images overall, as they tend to produce too few eligible pairs under the $100$ image constraint.

\subsection{SAE-based LatentDiff on COCO}
We use CLIP ViT-L/14 as the image encoder $\phi$ and the recently proposed monosemantic sparse autoencoder provided by ~\citet{pach2025sparseautoencoderslearnmonosemantic}. Without monosemanticity filtering, the highest-JSD features often align with broad concepts; we therefore apply a monosemanticity prior and retain only the top $50\%$ of neurons by a precomputed score

We assign concepts to neurons by using a sufficiently large vocabulary (in our case, 20k most common English words)\footnote{https://github.com/first20hours/google-10000-english/blob/master/20k.txt} and compare them to each SAE neuron's direction in CLIP's latent space. For each neuron, we pick the top-5-words with the highest cosine similarity to represent that neuron.

After the datasets have been passed through the SAE and JSD has determined which concepts are the most biased towards $D_A$ and $D_B$, we take the top-$k$ neurons ($k=5$) from each side and take the maximum cosine similarity between the ground truth and the proposed neuron(s) in the sentence-transformer latent space\footnote{https://github.com/huggingface/sentence-transformers}. We prefer sentence-transformers for text-to-text comparison over CLIP, as we find the contrastive training objective of CLIP yields subpar results in semantic similarity tasks.

\begin{table}[t]
  \centering
  \caption{Full results for \texttt{Noisy-Diff} (COCO) on \texttt{LatentDiff}. Cosine similarity is averaged over pairs; $\rho (\text{scarce})$ is the Pearson correlation between performance and scarcity.}
  \label{tab:full_results}
  \begin{tabular}{l r c c}
    \toprule
    Parent & \# Exper. & Cosine sim & $\rho$(scarce) \\
    \midrule
    bicycle     & 231 & 0.547 $\pm$ 0.182 & -0.01\\
    backpack    & 561 & 0.634 $\pm$ 0.195 &  0.03\\
    cat         & 253 & 0.607 $\pm$ 0.197 & -0.07\\
    dog         & 325 & 0.552 $\pm$ 0.175 &  0.01\\
    tv          & 378 & 0.527 $\pm$ 0.115&  0.03\\
    motorcycle  &  78 & 0.513 $\pm$ 0.144 &  0.13\\
    \midrule
    \textbf{Average} & \textbf{304} & \textbf{0.563} & \textbf{0.051}\\
    \bottomrule
  \end{tabular}
\end{table}
We evaluate retrieval quality by computing, for each side of a pair, the cosine similarity between the ground-truth COCO label for the missing mode (e.g., A) and the best-aligned concept among the top-5 JSD-selected SAE neurons. \Cref{tab:comparison_results} presents the results of a comparison study between our method and VisDiff~\cite{dunlap2024visdiff} on \texttt{Noisy-Diff} (COCO). Overall, we observe that VisDiff exhibits lower and more variable retrieval performance on rare missing modes, as reflected by both reduced cosine similarities and larger standard deviations. The behavior appears consistent with VisDiff's reliance on random sampling and captioning, which can occasionally recover the correct mode but may also fail to capture infrequent concepts depending on the sampled subset.
For this comparison, we restrict to 15 experiments per parent category, reflecting practical constraints in applying VisDiff at larger scales, as its computational cost grows rapidly with dataset size. Specifically, under these settings, VisDiff incurs $5.4$ hours of runtime, while \texttt{LatentDiff} takes $0.96$ hours (with no precomputed embeddings) on equivalent hardware.

We further analyze how performance varies with the scarcity of the missing modes. Across parent categories, Pearson correlations indicate no reliable association between rarity and \texttt{LatentDiff}'s ability to surface the missing mode. In contrast, VisDiff's performance degrades noticeably as missing modes become rarer, which is consistent with the sensitivity of its sampling-based procedure.

\Cref{tab:full_results} presents the full results for our instantiation of \texttt{Noisy-Diff} (COCO), across all possible $(A,B)$ for a given parent, as outlined in \Cref{subsec:dataset}. While the overall trends align with those observed in \Cref{tab:comparison_results}, these results more clearly demonstrate the scalability and consistency of the proposed approach.

Lastly, we note a qualitative limitation of our evaluation data. By construction, the \texttt{Noisy-Diff} (COCO) ground-truth labels are object categories that differ between $D_A$ and $D_B$. In practice, however, salient differences often emerge that are not captured by those labels. E.g., in the experiment $(P,A,B) = $ (Dog, Horse, Computer), \texttt{LatentDiff} gives ``Pasture'' (cosine sim $= 0.58$), which plausibly distinguishes the sets (horses frequently occur in pastures; computers do not) even though pasture is not our annotated ground truth. Synonym expansion marginally helps, but cannot fully resolve the context-dependency. We therefore view our quantitative scores as conservative, and complement them with DRE image-surfacing and qualitative examples.

\subsection{DRE based LatentDiff on COCO}
Complementary to our SAE-based pipeline, we observe that applying DRE to retrieve high-contrast images enables recovery of missing modes in several cases where the SAE pipeline is limited by the absence of neurons corresponding to a particular concept. Moreover, this approach can be viewed as an extension of VisDiff that replaces repeated random sampling with a more targeted and data-efficient selection of informative images from $D_A$ and $D_B$. 

Concretely, we train a lightweight DRE model to retrieve the top-10 high-contrast images from each dataset, caption these images using BLIP-2 ~\cite{li2017blip2}, and then prompt GPT-5-mini to describe the differences between the two resulting caption sets. Importantly, the dominant computational overhead of this procedure arises from embedding images into CLIP's latent space, rather than training the DRE model itself, which typically converges in under 15 seconds (on all \texttt{Noisy-Diff} (COCO) splits). As this method is generally complementary to our SAE-based method, caching CLIP embeddings results in an incredibly efficient pipeline, with the limiting factors being BLIP-2 and GPT-5-mini.

\begin{figure}[!tp]
    \centering
    \includegraphics[width=\linewidth]{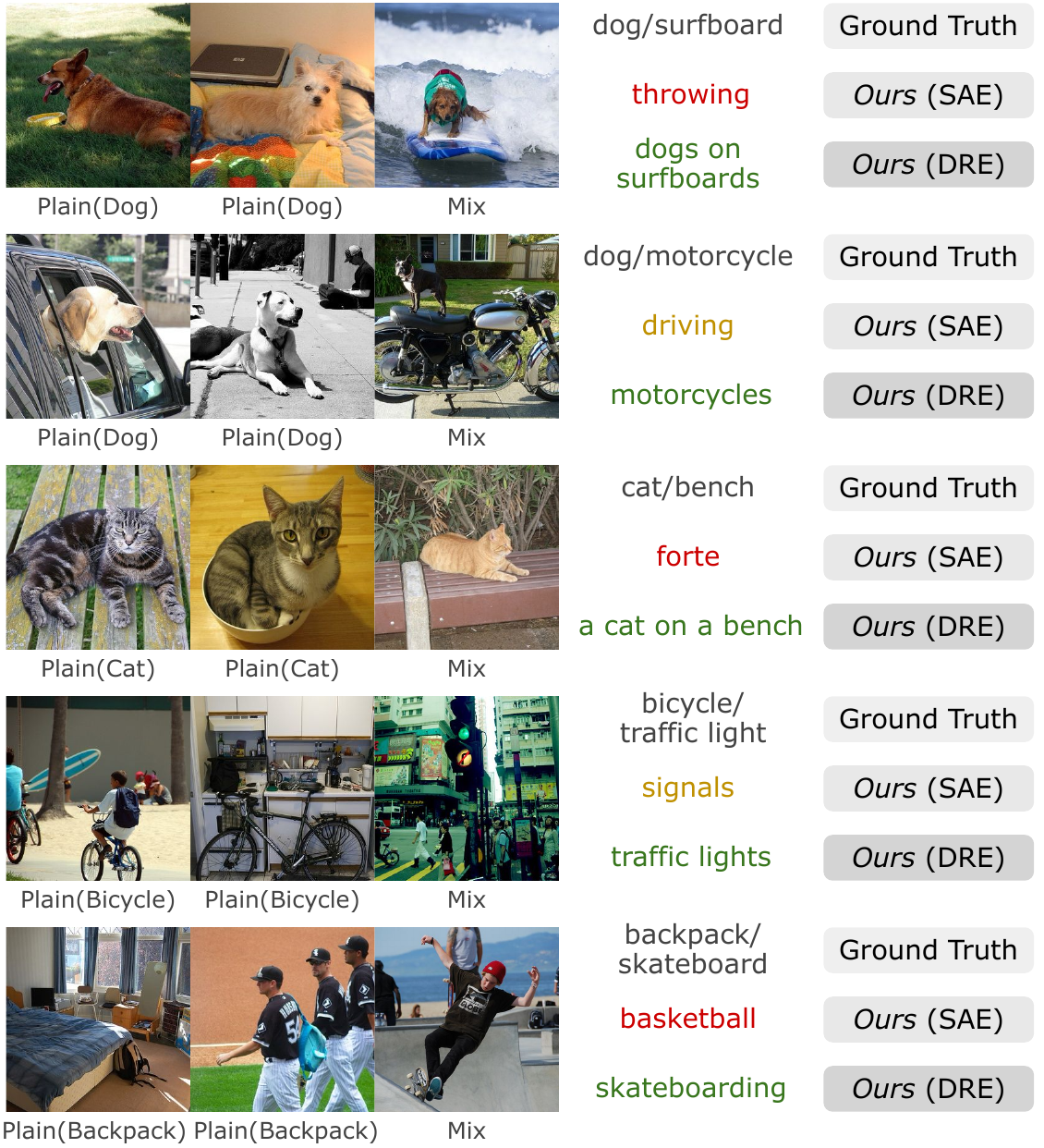}
    \caption{Failure cases of \texttt{LatentDiff} (SAE) where \texttt{LatentDiff} (DRE) is successful. We only show either $D_A$ or $D_B$ for simplicity. 
    \textcolor[HTML]{38761d}{Green} indicates success cases, \textcolor[HTML]{bf9000}{yellow} indicates partial success, and \textcolor[HTML]{cc0000}{red} indicates failure cases.}
    \label{fig:jsdvsdre}
\end{figure}

Running the same experiments as those reported in \Cref{tab:comparison_results}, we observe that this DRE-based procedure improves upon VisDiff, achieving a cosine similarity of $0.468 \pm 0.299$ and a substantially reduced dependency on abundance of the missing mode, with $\rho$(scarce)$= 0.018$. Although this approach does not match the performance of the JSD-based methodology overall, it proves effective as a complementary methodology when JSD fails. In particular, SAE-based methods may lack neurons corresponding to certain concepts, leading to low similarity scores due to the inability to assign an appropriate label. In several cases in Fig.\ \ref{fig:jsdvsdre}, DRE successfully retrieves images that strongly align with the missing concept, and the subsequent captioning and prompting pipeline is able to recover the corresponding mode. 

\subsection{Ablation Analysis}
\begin{figure}
    \centering
    \includegraphics[width=\linewidth]{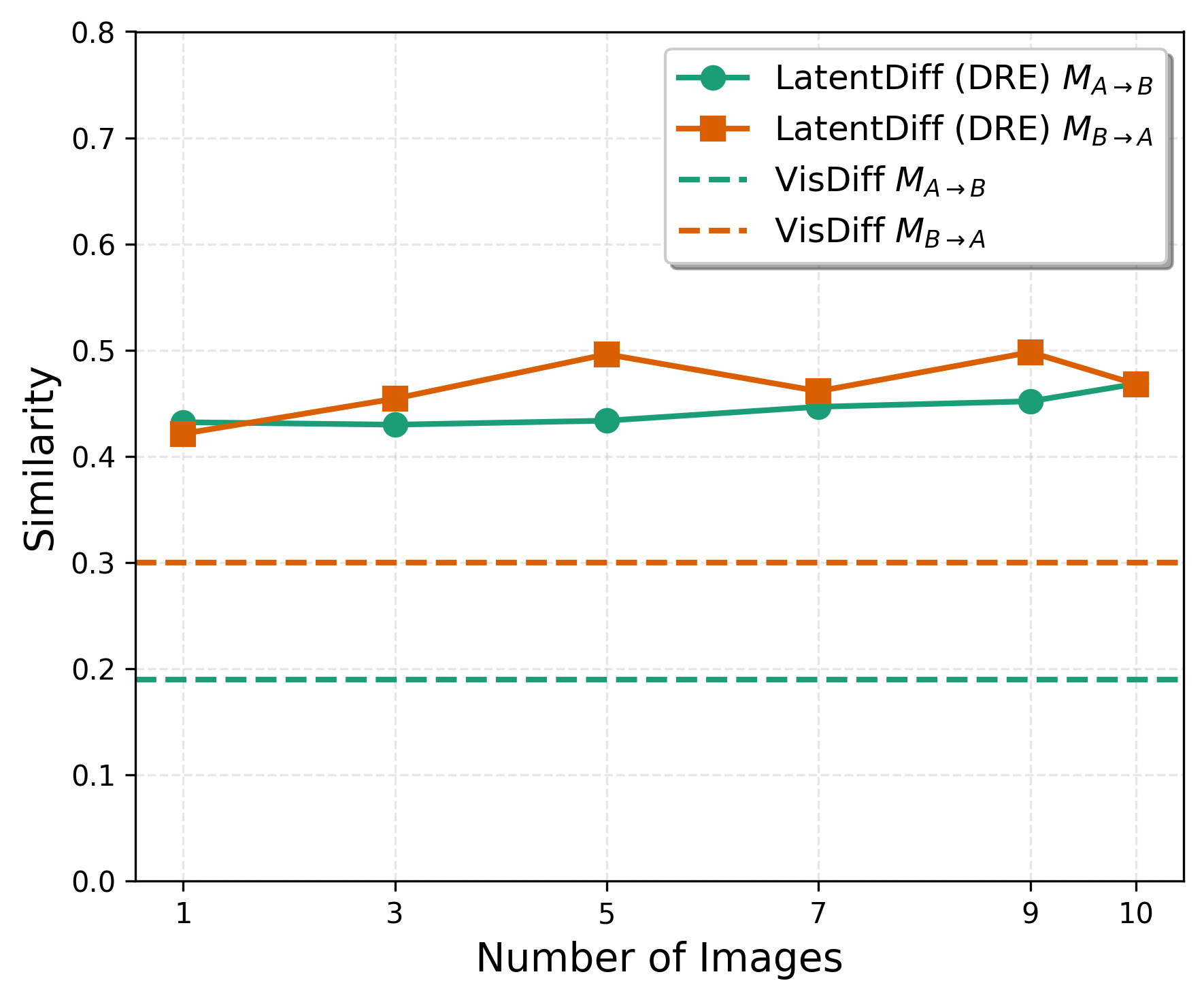}
\caption{Top-1 semantic similarity versus the number of images retrieved for missing-mode directions $M_{A \to B}$ and $M_{B \to A}$.
\texttt{LatentDiff} outperforms VisDiff across all retrieval budgets.}
    \label{fig:dre-sensitvity}
\end{figure}

\noindent \textbf{Sensitivity of LatentDiff (DRE) to the Number of Retrieved Samples.}
We study how the performance of the DRE-based localization depends on the number of top-ranked images retrieved from each dataset for downstream captioning and evaluation.
Figure~\ref{fig:dre-sensitvity} reports Top-1 semantic similarity as a function of the number of DRE-selected images for both missing-mode directions, $ M_{A \to B}$ and $M_{B \to A}$, and includes VisDiff as a reference baseline. We observe that performance is already stable when retrieving as few as 3-5 images, suggesting that density-ratio ordering is effective at surfacing highly discriminative samples even under sparse shifts. Further, increasing the retrieval budget yields only modest improvements. Interestingly, \texttt{LatentDiff} (DRE) consistently outperforms the VisDiff baseline (computed using $100$ images)  across all retrieval budgets. 
Overall, this ablation supports our design choice of using DRE as a lightweight localization mechanism: a small number of targeted samples is sufficient to guide downstream captioning and language-based reasoning, achieving improved accuracy at substantially lower computational cost.

We provide further sensitivity analyses to hyperparameters in Appendix~\ref{apx:sensitivity}.

\subsection{Joint Coverage Analysis}
\label{sec:joint}
While we introduce two different methodologies for recovering missing modes, \texttt{LatentDiff} (SAE) and \texttt{LatentDiff} (DRE), it is essential to demonstrate that these methods are complementary and that using them jointly provides measurable benefits over either approach in isolation. Both methods operate on the same underlying latent image representations, allowing embeddings to be cached per experiment; as a result, running the two procedures sequentially incurs minimal computational overhead.

Although our qualitative analysis shows that DRE can capture semantic modes that are not well represented by the SAE (\Cref{fig:jsdvsdre}), DRE alone performs less favorably on objective retrieval metrics. This motivates a combined evaluation. In this section, we analyze joint coverage, comparing VisDiff, \texttt{LatentDiff} (SAE), \texttt{LatentDiff} (DRE), and \texttt{LatentDiff} (Combined). The combined variant selects candidates from both SAE-derived neuron activations and DRE-generated captions based on confidence (top-k).

\begin{figure}
    \centering
    \includegraphics[width=\linewidth]{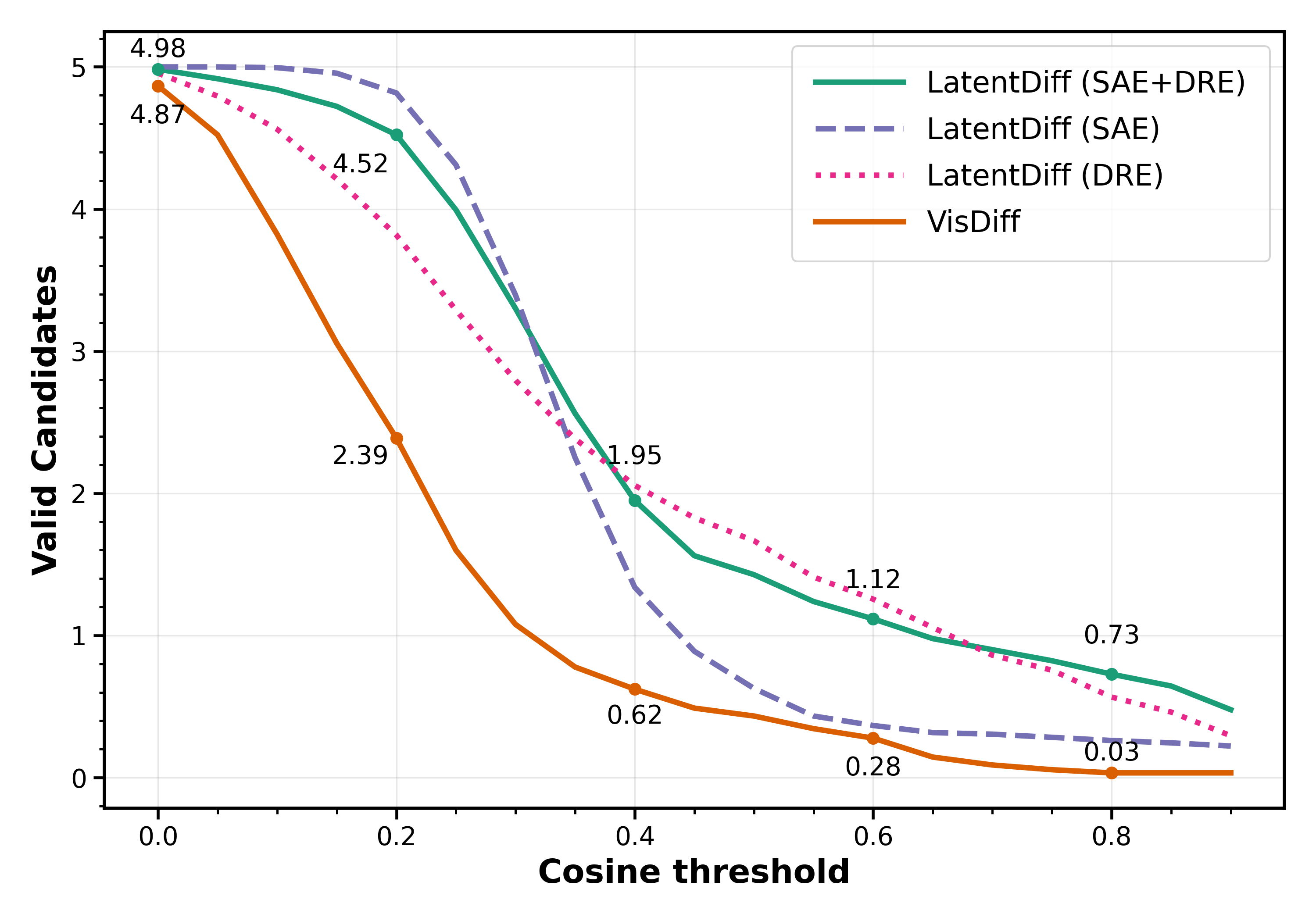}
\caption{Joint coverage of missing modes as a function of cosine similarity threshold. We report the expected number of aligned candidates among the top-5 proposals for \texttt{LatentDiff} (SAE), \texttt{LatentDiff} (DRE), and their combination (SAE+DRE; top-3 SAE neurons + top-2 DRE captions), and VisDiff.}
    \label{fig:joint-coverage}
\end{figure}

Our results in \Cref{fig:joint-coverage} show that aggregating candidates from both sources consistently yields stronger coverage than either SAE or DRE alone. Importantly, in practical settings, practicioners cannot know \textit{a priori} whether the latent concept space exposed by the SAE or the DRE will be more effective for a given task. By leveraging both mechanisms, the combined approach reduces the risk of selecting an underperforming method and instead achieves more robust and reliable performance.

Further, we report that under the experimental settings shown in \Cref{tab:comparison_results}, the optimal combination of $3$ SAE $+$ $2$ DRE achieves an average cosine similarity of $0.678 \pm 0.196$, indicating a $23\%$ increase when these methods are used jointly, with the full results contained in Appendix~\ref{apx:joint}.

\subsection{Semantic Comparison at the Million-Image Scale}
\label{sec:imagenet}
In this section, we examine the application of \texttt{LatentDiff} to substantially larger datasets. We motivate this analysis by revisiting the two central advantages of \texttt{LatentDiff}: its ability to scale to larger and noisier datasets, and its capacity to surface rare missing modes. While prior experiments demonstrate the method's effectiveness in controlled settings (e.g., COCO parent categories) where most images share a common attribute, it remains unclear whether these benefits persist when dataset size increases by orders of magnitude.

We demonstrate this capability through an experiment on the ImageNet training set using the \texttt{Noisy-Diff} protocol, which contains over one million images; more than two orders of magnitude larger than COCO and substantially beyond the dataset scales explored in prior work (e.g., VisDiff). This experiment evaluates whether \texttt{LatentDiff} continues to function in regimes where the signal of interest is extremely sparse relative to the dataset size. Specifically, we partition the training set into two halves and remove a distinct WordNet class from each subset, then test whether \texttt{LatentDiff} can recover the missing mode between the resulting datasets. Unlike our COCO experiments, the dataset does not share a unifying parent category, and the removed classes may constitute as little as $1\%$ of the total data, producing a challenging setting in which the missing concept is heavily diluted.

Despite operating at the million-image scale, \texttt{LatentDiff} remains both effective and computationally practical. In contrast to approaches such as VisDiff, which rely on repeatedly sampling and prompting/captioning images, the cost of \texttt{LatentDiff} is dominated by a single embedding pass over the dataset. For this experiment, the embedding stage takes approximately 10 hours on a single NVIDIA H100 (the current implementation does not contain any optimization for speed, i.e., exploiting parallelism, and a more efficient data loader), after which dataset comparisons operate purely in latent space and are complete in only a few minutes. Consequently, the method scales linearly with dataset size and avoids the repeated inference costs inherent in sampling-based approaches. Empirically, we obtain an overall cosine similarity of $0.41$, with representative examples including the recover of concepts corresponding to \textit{Building} (neuron concept: Builder), \textit{Establishment} (neuron concept: Apartment), and \textit{Protective Cover} (neuron concept: Armor). 

\begin{figure}
    \centering
    \includegraphics[width=\linewidth]{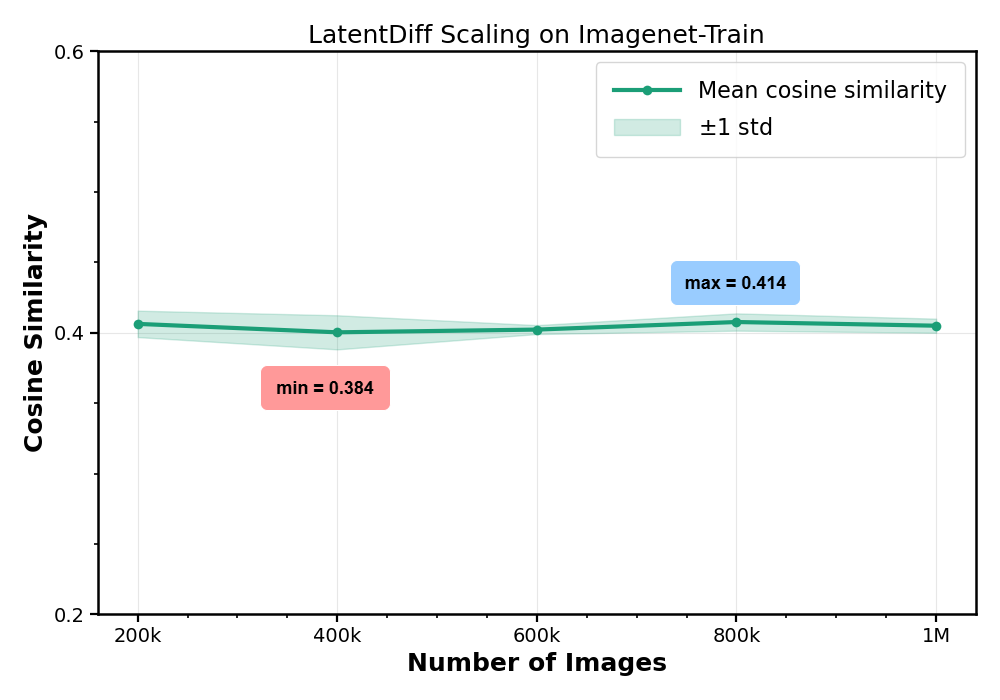}
\caption{Scaling experiment on the ImageNet training set. We evaluate \texttt{LatentDiff} on progressively larger subsets of the dataset (200k–1M images). The mean cosine similarity between predicted and ground-truth missing modes remains stable across scales, with the shaded region indicating $\pm1$ standard deviation across runs.}
    \label{fig:scaling-exp}
\end{figure}

Further, \Cref{fig:scaling-exp} strengthens our scaling claim by showing that the mean cosine similarity of \texttt{LatentDiff} remains largely unchanged across progressively larger ImageNet subsets. Despite increasing the dataset size from a few hundred thousand images to the full million, performance remains stable, with very minor variance across experiments. This indicates that \texttt{LatentDiff} remains reliable with respect to scale, and its effectiveness is not tied to a particular methodology that degrades as the datasets increase in size. This demonstrates that \texttt{LatentDiff} can be applied to dataset scales far beyond those previously considered in semantic dataset comparison. Additional methodological details are provided in \Cref{apx:imagenet}.

\section{Discussion and Future Works}
Our experiments reveal several important insights about semantic dataset comparison. First, operating in pretrained latent spaces provides a strong foundation for detecting distributional differences without expensive per-image processing. Second, the complementary nature of SAE-based and DRE-based analysis proves crucial. While SAEs offer efficient, interpretable concept-level comparisons, they may fail to learn disentangled representations for certain concepts. DRE compensates by enabling open-vocabulary discovery through targeted captioning. This hybrid strategy consistently outperforms either approach alone, particularly when differences are rare or involve concepts not well-represented by SAE neurons.
A key limitation of our approach is its dependence on SAE quality. As sparse autoencoders for vision models continue to improve in coverage and monosemanticity, we expect corresponding gains in their performance. Additionally, while our evaluation focuses on object-level differences, extending to more abstract attributes (e.g., style, composition) remains an open challenge.

Several promising avenues emerge from this work. First, training task-specific SAEs on domain-relevant data (e.g., medical imaging, satellite imagery) could improve concept coverage for specialized applications. Second, incorporating hierarchical SAE could enable multi-granularity comparisons, surfacing both coarse category-level and fine-grained attribute-level differences simultaneously. 
Finally, integrating human-in-the-loop feedback to iteratively refine discovered differences would enhance practical utility.





\section*{Acknowledgements}
This work was performed under the auspices of the U.S. Department of Energy by Lawrence Livermore National Laboratory under Contract DE-AC52-07NA27344. The work is partially funded by LLNL LDRD 23-ERD-029, DOE ECRP 51917/SCW1885. This manuscript is reviewed and released under LLNL-PROC-2015604.


\newpage
\appendix
\onecolumn

\section{Sparse Autoencoder and Density Ratio Estimation Definition}
\label{apx:background}

In this work, we leverage a sparse autoencoder to infuse semantics to otherwise opaque latent spaces and use density ratio estimation in the latent space for identifying the most distinguishing samples. A brief description of both methods is discussed below.



\subsection{Sparse Autoencoder}
\label{appendix:SAE}
The general purpose of the Sparse Autoencoder is to find a sparse reinterpretation of the latent space representation of a given deep learning architecture. Specifically, an SAE implements a form of sparse dictionary learning, in which a signal is decomposed into a sparse combination of atoms from an overcomplete dictionary. Concretely, given a latent vector $h \in \mathbb{R}^d$, the SAE encodes $h$ into a sparse representation $z \in \mathbb{R}^k$ and reconstructs it via a learned dictionary $W \in \mathbb{R}^{d \times k}$, with $k = \epsilon d$ for some expansion factor $\epsilon \gg 1$.

The SAE consists of an encoder-decoder pair
\[
z = \sigma(W_{enc}^T(h-b)), \qquad \hat{h} = W_{dec}^Tz + b
\]

Where $W_{enc}$, $W_{dec} \in \mathbb{R}^{d \times k}$, $b \in \mathbb{R}^d$ is a shared bias, and $\sigma: \mathbb{R}^k \rightarrow \mathbb{R}^k$ is usually TopK. The parameters are trained to minimized a reconstruction-sparsity tradeoff

\[
\mathcal{L}(h) = ||h - \hat{h}||_2^2 + \lambda||z||_1
\]
where $\lambda$ controls the degree of sparsity. This sparse reinterpretation of the latent space is said to yield more interpretable, often monosemantic features, making it easier to attribute model behavior \cite{pach2025sparseautoencoderslearnmonosemantic}.

\subsection{Density Ratio Estimation}
\label{subsec:dre_bg}
Complementary to the sparse interpretations that SAE is intended for, Density Ratio Estimation (DRE) attempts to estimate the ratio between two probability densities using samples. In the domain of images, the DRE would rank individual images by how much they "look like" dataset $A$ vs.\ $B$. Suppose $p_A$ and $p_B$ are the latent distributions of their corresponding datasets, and the target is the density ratio,
\[
r(h) = \frac{p_A(h)}{p_B(h)}
\]
When the ratio is low, it is more likely that $h$ is from distribution $p_A$ and out of distribution for $p_B$. Practically, density ratio estimation is implemented as a probabilistic classifier $f_{\theta}(h)$ trained to distinguish images of $A$ (label 1) from $B$ (label 0).



\section{VisDiff Benchmark}
\label{apx:visdiff}
We additionally evaluate our method on the VisDiff benchmark dataset ~\cite{dunlap2024visdiff}, which consists of Paired Image Sets (PIS) that are intentionally constructed to be semantically pure. This contrasts with our own benchmark, which is designed to include substantial noise and rarity of missing mode. The corresponding results are reported in \Cref{tab:pis_visdiff_latentdiff}, though they should be interpreted with appropriate caution.
\begin{table}[!htbp]
\centering
\caption{Cosine similarity on PIS difficulty splits (mean $\pm$ std). For VisDiff, BLIP-2 as the captioner, GPT-5-Mini as the proposer, CLIP as the ranker, and GPT-5.1 as the evaluator}
\label{tab:pis_visdiff_latentdiff}
\begin{tabular}{lcc}
\toprule
 & \textbf{VisDiff} & \textbf{LatentDiff (SAE)} \\
\midrule
\textbf{PIS Easy}   & $0.70 \pm 0.12$ & $0.82 \pm 0.18$ \\
\textbf{PIS Medium} & $0.65 \pm 0.16$ & $0.66 \pm 0.21$ \\
\textbf{PIS Hard}   & $0.52 \pm 0.25$ & $0.55 \pm 0.20$ \\
\bottomrule
\end{tabular}
\end{table}
In this benchmark, difficulty is defined in terms of the subtlety of the semantic distinction between the paired image sets. For example, one of the hardest pairs contrasts \emph{“lace wedding dresses”} with \emph{“satin wedding dresses”}, for which the underlying difference is \emph{lace vs.\ satin}. While correctly recovering the missing attribute in such cases naturally yields higher cosine similarity, a method may also achieve a non-trivial score by merely restating the shared parent concept (e.g., \emph{wedding dresses}).

To mitigate this effect, we apply a pruning strategy to the SAE representations when conducting experiments (both on this benchmark, and globally). Specifically, we remove globally dominant neurons that exhibit high activation across both datasets in a pair, as these neurons typically correspond to shared parent concepts rather than the distinguishing attribute. In practice, we discard the top $10\%$ of most active neurons under the joint distribution ${D_A, D_B}$, ensuring that concepts such as \emph{wedding dress} do not artificially inflate similarity scores and that performance reflects sensitivity to the intended semantic distinction.

We use a Sentence-Transformers model~\cite{reimers-2019-sentence-bert} to compute all cosine similarity scores reported in our experiments (both in this section, and in all experiments). This choice is motivated by two considerations. First, we aim to avoid biasing the evaluation toward the latent space of any text encoder used within the compared methods themselves. By using an external semantic embedding model that is not part of either pipeline, we obtain a more neutral similarity metric. Second, Sentence-Transformers are explicitly trained to produce semantically meaningful sentence and phrase embeddings under cosine similarity, making them well suited for comparing short descriptive captions and concept phrases like what VisDiff and our own \texttt{LatentDiff} (DRE) gives.

In contrast, cosine similarity applied to standard token embeddings or non–sentence-level representations is often unreliable for multi-word expressions, since such embeddings are not optimized for compositional semantic similarity. Sentence-Transformers address this by training with contrastive and similarity-based objectives so that cosine distance directly reflects semantic relatedness at the phrase and sentence level.

\paragraph{Beyond Object-level Differences}

We provide a brief qualitative analysis on PIS-hard to illustrate the types of semantic shifts captured by \texttt{LatentDiff}. As a reminder, PIS-hard is LLM-assisted and human-curated (as introduced in VisDiff). Table below shows representative examples recovered by the SAE branch.

\begin{center}
\begin{tabular}{lll}
\toprule
Ground-truth Shift & Recovered Concept & Shift Type \\
\midrule
spotted vs.\ striped butterflies & spots / striped & texture / pattern \\
electric vs.\ diesel trucks & electrical & functional \\
Rolex vs.\ Omega watches & rolex / omega & brand \\
playing chess vs.\ checkers & checker & activity \\
porcelain vs.\ ceramic plates & porcelain / ceramic & material \\
baroque vs.\ classical orchestras & baroque & genre \\
WWI vs.\ WWII uniforms & ww2 & historical context \\
medieval Europe vs.\ ancient Rome films & medieval / rome & historical setting \\
gothic vs.\ modernist architecture & gothic & style \\
Art Deco vs.\ Brutalist buildings & deco & style \\
\bottomrule
\end{tabular}
\end{center}

These examples demonstrate that \texttt{LatentDiff} is able to recover a range of semantic differences beyond simple object identity, including texture, material, activity, brand, and historical context. Notably, many of these shifts are abstract (e.g., style or historical setting) rather than tied to discrete objects, suggesting that the learned latent features capture higher-level structure in the data.

We emphasize that this analysis is preliminary and qualitative. While these results provide evidence that the method extends beyond object-level differences, more systematic evaluation of complex semantic shifts (e.g., artistic style, scene composition, or multi-object interactions) remains an important direction for future work.

\section{GSCLIP Benchmark}
\label{apx:gsclip}

We additionally evaluate against GSCLIP~\cite{zhu2022gsclip} as a representative baseline for latent-driven explanation methods, in contrast to VisDiff which is caption-based. We emphasize, however, that GSCLIP and \texttt{LatentDiff} are designed for substantially different problem formulations. GSCLIP operates in a \emph{closed-set hypothesis selection} setting, where the task is to select an explanation from a relatively small candidate set, whereas \texttt{LatentDiff} is designed for \emph{open-vocabulary semantic dataset comparison}, where the goal is to discover differences without assuming a predefined answer space. As such, the comparison is not perfectly aligned; nevertheless, we construct a best-effort, one-to-one evaluation in both directions.

\paragraph{GSCLIP on NoisyDiff.}
To adapt GSCLIP to our setting without introducing oracle bias, we do not provide the ground-truth answer space as candidate explanations. Instead, we follow GSCLIP's language-model-based generator protocol, instantiated over the same open vocabulary used by our method (the \texttt{clip\_disect\_20k} vocabulary). We further employ the same negation-based formulation used in GSCLIP to derive differences between datasets, and run the procedure in both directions to obtain explanations for each dataset. We note that the original pairwise formulation in GSCLIP becomes computationally infeasible under a 20k vocabulary, and thus this adaptation is necessary for scalability.

Under this setting (identical to \Cref{tab:comparison_results}), GSCLIP achieves a score of \textbf{0.47}, compared to \textbf{0.55} for our SAE-only branch and \textbf{0.68} for the full SAE+DRE \texttt{LatentDiff} pipeline.

To ensure that our results are not driven by cosine similarity in embedding space, we also evaluate both methods using GSCLIP's own top-$k$ accuracy metric:

\begin{center}
\begin{tabular}{lcccc}
\toprule
Method & Top-1 Exact & Top-5 Exact & Top-1 Synonym & Top-5 Synonym \\
\midrule
GSCLIP & 2.78\% & 12.2\% & 2.78\% & 12.78\% \\
LatentDiff & \textbf{7.78\%} & \textbf{23.9\%} & \textbf{11.1\%} & \textbf{24.4\%} \\
\bottomrule
\end{tabular}
\end{center}

We note that these accuracies are lower than those reported in~\cite{zhu2022gsclip}, which is expected given the significantly more challenging open-vocabulary setting. Here, both methods must recover correct concepts from a much larger candidate space, making exact word (or WordNet synonym) recovery substantially less likely.

\paragraph{LatentDiff on MetaShift.}
We additionally evaluate \texttt{LatentDiff} in GSCLIP's native setting using the MetaShift dataset. To ensure a fair comparison, we restrict our method to the same answer space used by GSCLIP (66 candidate labels), effectively converting our approach into a $k$-class selection problem. Following GSCLIP, a prediction is considered correct if it recovers \emph{either} of the two ground-truth labels associated with the dataset pair.

On a random sample of 100 dataset pairs (following the GSCLIP evaluation protocol), we obtain:

\begin{center}
\begin{tabular}{lccc}
\toprule
Method & Top-1 & Top-3 & Top-5 \\
\midrule
GSCLIP & 30\% & 50\% & 63\% \\
LatentDiff & \textbf{46\%} & \textbf{62\%} & \textbf{72\%} \\
\bottomrule
\end{tabular}
\end{center}

\paragraph{Discussion}
These experiments demonstrate that \texttt{LatentDiff} remains competitive under both its native open-vocabulary evaluation and GSCLIP-style closed-set metrics. Moreover, when evaluated within GSCLIP's preferred setting, \texttt{LatentDiff} outperforms GSCLIP across all top-$k$ thresholds. At the same time, we emphasize that the two methods target different regimes: GSCLIP is most effective when selecting from small, curated candidate sets, whereas \texttt{LatentDiff} is designed for scalable, open-ended semantic difference discovery. We therefore view this comparison as complementary rather than definitive, and as supplementary evidence that \texttt{LatentDiff} is not pre-biased by the cosine similarity of the sentence-transformer.

\section{Joint Coverage}
\label{apx:joint}

We provide the full joint coverage table as mentioned in \Cref{sec:joint} across all SAE+DRE ensemble combinations for the experiment outlined in \Cref{tab:comparison_results}. 

\begin{center}
\begin{tabular}{lcc}
\toprule
Ensemble & Avg & Std \\
\midrule
5 SAE + 0 DRE & 0.548 & 0.163 \\
4 SAE + 1 DRE & 0.642 & 0.148 \\
3 SAE + 2 DRE & \textbf{0.678} & 0.196 \\
2 SAE + 3 DRE & 0.674 & 0.203 \\
1 SAE + 4 DRE & 0.651 & 0.205 \\
0 SAE + 5 DRE & 0.615 & 0.203 \\
\bottomrule
\end{tabular}
\end{center}

As shown above, combining SAE and DRE components consistently improves performance over either method in isolation. In particular, mixed ensembles outperform the pure SAE (5+0) and pure DRE (0+5) configurations, indicating that the two components capture complementary signals. Performance peaks at the 3 SAE + 2 DRE setting, achieving a 23.7\% relative improvement over the SAE-only baseline, with nearby configurations (e.g., 2 SAE + 3 DRE) exhibiting similar performance. This suggests that the method is robust to the exact mixture used and does not rely on a narrow tuning.

\section{NoisyDiff with ImageNet}
\label{apx:imagenet}
As discussed in \Cref{sec:imagenet}, we use the ImageNet training set to conduct further testing on our \texttt{NoisyDiff} protocol and \texttt{LatentDiff} methodology. The standard ImageNet validation set contains only 1000-1300 images per class, which is insufficient for controlled missing-mode experiments at the level of individual categories. To address this limitation, we construct larger, semantically coherent supersets of ImageNet classes by leveraging the WordNet hierarchy underlying ImageNet labels.

Each ImageNet class corresponds to a WordNet synset. We group the ImageNet synsets into coarser semantic categories by mapping each synset to a hypernym ancestor at a fixed depth in the WordNet graph. To ensure coverage, we select the most specific ancestor whose depth does not exceed the target cut depth. This procedure yields on the order of 100 semantic groups (e.g., dog, building, clothing), with each aggregating images from multiple closely-related ImageNet classes.

Using these WordNet-derived groups, we follow the \texttt{NoisyDiff} methodology outlined in \Cref{sec:noisediff}. This results in larger splits that are $\approx$ 650,000 images per side and differ only in the presence or absence of a specific semantic mode, ranging anywhere from 6,500 to 30,000 images.

\section{Sensitivity Analysis}
\label{apx:sensitivity}

We evaluate the sensitivity of \texttt{LatentDiff} to key design and training hyperparameters. Overall, we find that performance remains stable across a broad range of settings, with degradation occurring only under extreme pruning.

\paragraph{Pipeline hyperparameters.}
We first analyze sensitivity to three components of the \texttt{LatentDiff} pipeline: (i) the number of vocabulary entries retained per neuron (top-$k$), (ii) the monosemanticity filtering threshold, and (iii) cosine-similarity-based pruning of neurons. Results are reported using the same evaluation protocol as Table~1.

\begin{center}
\begin{tabular}{lccc}
\toprule
Setting & Value & Avg & Std \\
\midrule
top-$k$ & 3  & 0.550 & 0.171 \\
top-$k$ & 5  & 0.565 & 0.163 \\
top-$k$ & 7  & 0.567 & 0.166 \\
top-$k$ & 10 & 0.561 & 0.163 \\
\midrule
monosem & 0.25 & 0.419 & 0.097 \\
monosem & 0.50 & 0.565 & 0.158 \\
monosem & 0.75 & 0.594 & 0.178 \\
monosem & 1.00 & 0.573 & 0.189 \\
\midrule
cos prune & 0.05 & 0.576 & 0.171 \\
cos prune & 0.10 & 0.579 & 0.172 \\
cos prune & 0.15 & 0.598 & 0.172 \\
cos prune & 0.20 & 0.592 & 0.166 \\
\bottomrule
\end{tabular}
\end{center}

Across these settings, performance is largely stable. In particular, the method is relatively insensitive to the choice of top-$k$ within a reasonable range. Monosemanticity filtering improves performance up to a moderate threshold, while overly aggressive filtering (e.g., 0.25) significantly degrades results by discarding useful features. Cosine-based pruning exhibits similar robustness, with only modest variation across thresholds.

\paragraph{SAE training hyperparameters.}
We additionally analyze sensitivity to the underlying sparse autoencoder (SAE) training configuration, including the expansion factor and the BatchTop-$k$ sparsity parameter used during training.

\begin{center}
\begin{tabular}{lcc}
\toprule
SAE Setting & Avg & Std \\
\midrule
Expansion $\times 2$  & 0.535 & 0.181 \\
Expansion $\times 4$  & 0.511 & 0.164 \\
Expansion $\times 8$  & 0.521 & 0.165 \\
Expansion $\times 16$ & 0.509 & 0.147 \\
Expansion $\times 64$ & 0.548 & 0.133 \\
\midrule
top-$k$ = 5  & 0.526 & 0.169 \\
top-$k$ = 20 & 0.548 & 0.133 \\
top-$k$ = 30 & 0.613 & 0.186 \\
top-$k$ = 50 & 0.612 & 0.181 \\
\bottomrule
\end{tabular}
\end{center}

These results indicate that \texttt{LatentDiff} is also robust to reasonable variations in SAE architecture and training sparsity. While performance improves with moderately larger top-$k$ values, the method does not rely on a narrowly tuned configuration. We initialize top-$k=20$ in our own experiments following \cite{pach2025sparseautoencoderslearnmonosemantic}, who fix the BatchTopK activation at this value and note that it provides a reasonable tradeoff between interpretability and reconstruction quality.

Taken together, these results suggest that \texttt{LatentDiff} is not highly sensitive to the specific heuristic or training choices used in our implementation. Instead, the method exhibits stable performance across a wide range of configurations, supporting the robustness of the overall approach.

\end{document}